\documentclass[10pt,twocolumn,letterpaper]{article}

\usepackage{iccv}
\usepackage{times}
\usepackage{epsfig}

\usepackage{subcaption}
\usepackage{graphicx}
\usepackage{amsmath}
\usepackage{amssymb}
\usepackage{booktabs}
\usepackage{tabularx}
\usepackage{dsfont}
\usepackage[export]{adjustbox}
\usepackage{array,multirow,graphicx}
\usepackage{pgfplotstable}
\usepackage{pgfplots}
\usetikzlibrary{shapes.geometric}
\pgfplotsset{compat=newest}
\pgfdeclareplotmark{mystar}{
    \node[star,star point ratio=2.25,minimum size=6pt,
          inner sep=0pt,draw=black,solid,fill=red] {};
}

\usepackage[pagebackref=true,breaklinks=true,letterpaper=true,colorlinks,bookmarks=false]{hyperref}
\usepackage[capitalize]{cleveref}
\crefname{section}{Sec.}{Secs.}
\Crefname{section}{Section}{Sections}
\Crefname{table}{Table}{Tables}
\crefname{table}{Tab.}{Tabs.}

 \iccvfinalcopy %

\def\iccvPaperID{11828} %

\ificcvfinal\fi

\begin{document}

\title{Discriminative Class Tokens for Text-to-Image Diffusion Models}

\author{Idan Schwartz\textsuperscript{\rm 1}\thanks{Equal contribution.} \quad Vésteinn Snæbjarnarson\textsuperscript{\rm 2}\footnotemark[1] \quad  Sagie Benaim\textsuperscript{\rm 2}\quad  Hila Chefer\textsuperscript{\rm 1}\\
Ryan Cotterell\textsuperscript{\rm 3}\qquad  Lior Wolf\textsuperscript{\rm 1}\qquad  Serge Belongie\textsuperscript{\rm 2}\\
\textsuperscript{\rm 1}Tel Aviv University\quad \textsuperscript{\rm 2}University of Copenhagen \quad \textsuperscript{\rm 3}ETH Zürich}

\maketitle
\ificcvfinal\thispagestyle{empty}\fi

\begin{abstract}

Recent advances in text-to-image diffusion models have enabled the generation of diverse and high-quality images. However, generated images often fall short of depicting subtle details and are susceptible to errors due to ambiguity in the input text. One way of alleviating these issues is to train diffusion models on class-labeled datasets. This comes with a downside, doing so limits their expressive power: (i) supervised datasets are generally small compared to large-scale scraped text-image datasets on which text-to-image models are trained, and so the quality and diversity of generated images are severely affected, or (ii) the input is a hard-coded label, as opposed to free-form text, which limits the control over the generated images.

In this work, we propose a non-invasive fine-tuning technique that capitalizes on the expressive potential of free-form text while achieving high accuracy through discriminative signals from a pretrained classifier, which guides the generation. This is done by iteratively modifying the embedding of a single input token of a text-to-image diffusion model, using the classifier, by steering generated images toward a given target class. Our method is fast compared to prior fine-tuning methods and does not require a collection of in-class images or retraining of a noise-tolerant classifier. We evaluate our method extensively, showing that the generated images are: (i) more accurate and of higher quality than standard diffusion models, (ii) can be used to augment training data in a low-resource setting, and (iii) reveal information about the data used to train the guiding classifier. The code is available at \url{https://github.com/idansc/discriminative_class_tokens}.

\end{abstract}

\section{Introduction}
\label{sec:intro}
Text-to-image diffusion models~\cite{DBLP:journals/corr/abs-2112-10752, dhariwal2021diffusion} have shown remarkable success in creating diverse and high-quality imagery conditioned on input text. However, they fall short when the input text contains lexical ambiguity or when generating fine-grained details. For instance, a user might wish to render an image of an `iron' for clothes, but could instead be presented with an image of the elemental metal.

One way to alleviate these issues is to use a pretrained classifier to guide the denoising process. 
One such method mixes the score estimate of a diffusion model with the gradient of the log probability of a pre-trained classifier~\cite{dhariwal2021diffusion}. However, this approach has the downside of needing a classifier that works on both real and noisy data. It is also possible to condition diffusion on a class label~\cite{ho2022classifier}. Unfortunately, training the models on curated datasets prevents them from fully utilizing the expressive power of a diffusion model trained on web-scale image-text pairs.  %

\begin{figure*}[hpt!]
    \centering
    \includegraphics[width=1\linewidth]{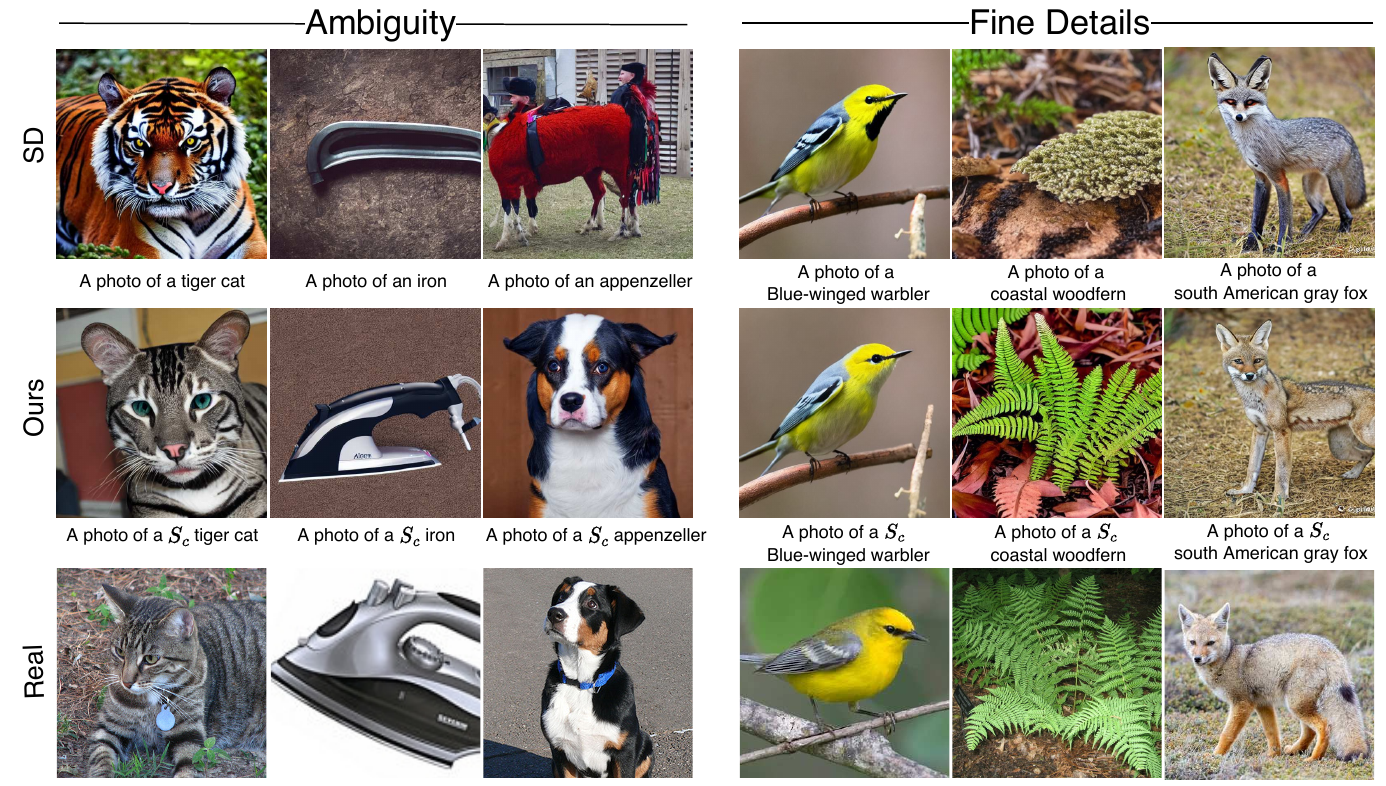}
    \vspace{-0.7cm}
    \caption{We propose a technique that introduces a token ($S_c$) corresponding to an external classifier label class $c$. This improves text-to-image alignment when there is lexical ambiguity and enhances the depiction of intricate details.}

    \label{fig:teas}
    \vspace{-0.3cm}
\end{figure*}

A different line of work 
fine-tunes a diffusion model, or some of its input tokens, using a small ({\raise.17ex\hbox{$\scriptstyle\sim$}}5) group of images~\cite{gal2022textual,kawar2022imagic, ruiz2022dreambooth}. %
These methods have the following drawbacks: (i) training new concepts is usually slow and takes upwards of a few hours, (ii) the result may change the generated images (as compared to the original diffusion model) to fit only the new label or concept, and (iii) generated images are based on features from a small group of images 
and may not capture the diversity of the entire class.

This work introduces a method that accurately captures the desired class, avoiding lexical ambiguity and accurately portraying fine-grained details. It does so while retaining the full expressive power of the pretrained diffusion model while avoiding the drawbacks mentioned above. 
Instead of guiding the diffusion process or updating the entire model with the classifier, we only update the representation of a single added token, one corresponding to each class of interest, without tuning the model on labeled images. When learning the token representation corresponding to a given target class, we iteratively generate new images with a higher class probability according to the pretrained classifier. 
At each iteration, feedback from the classifier steers the designated discriminative class token. Our optimization process uses a new technique, gradient skipping, which only propagates the gradient through the final stage of the diffu sion process. The optimized token can then be utilized to generate additional images using the original diffusion model. %

Our method has several advantages. First, unlike other class conditional methods such as \cite{dhariwal2021diffusion}, it only requires an off-the-shelf classifier and does not require a classifier trained on noisy data. Second, our method is fast and allows for ``plug-and-play" improvements of generated images by making use of a pre-trained token. This is in comparison to other methods, such as Textual Inversion~\cite{gal2022textual}, which can take a few hours to converge. 
Third, our method employs a classifier trained on an extensive collection of images without needing access to those images. This is beneficial as (i) the token is generated from the full set of class-discriminative features as opposed to features from a small set of images, and (ii) in some cases, such as when privacy concerns are involved, it is desirable to share only the classifier and not the data on which it is trained.

We evaluate our method both in fine-grained and coarse-grained settings. In the fine-grained setting, we investigate the ability of our method to generate details of species in the
CUB~\cite{WahCUB_200_2011} and iNat21~\cite{su2021semi_iNat} datasets. In the coarse setting, we consider the ImageNet~\cite{deng2009imagenet} dataset. Our primary metric is the accuracy of the generated samples as measured in two ways: (i) we show that our generated images are more often correctly classified using pre-trained classifiers, in comparison to baselines, and (ii) we show that classification models trained on generated samples, either on their own or in combination with a limited amount of real training data, result in improved accuracy.
We also measure the quality and diversity of the generated images compared to SD and another class-conditioned technique, showing that our method is superior in terms of the commonly used Fréchet inception distance (FID)~\cite{NIPS2017_8a1d6947}.  %
Finally, we include many qualitative examples demonstrating the effectiveness of our approach.
 In \cref{fig:teas}, we %
 resolve ambiguity in the input text and add discriminative features for a given class. In the ambiguous category, besides the \emph{iron} example, the image of a \emph{tiger cat} becomes the cat species instead of a tiger, and \emph{Appenzeller} moves from depicting a group of people, from the Appenzeller area to the dog species. In the fine-grained category, the bird's throat color is corrected, the shape features of the \emph{castal woodferm} are corrected, and the \emph{cresed morwong} show distinctive features that closely resemble those of the species.

\section{Related work}

The field of text-based image generation has been studied extensively, both for GANs and, more recently, for diffusion models~\cite{ding2021cogview,hinz2020semantic,tao2020df, li2019controllable,li2019object,qiao2019learn,qiao2019mirrorgan,ramesh2021zero,zhang2018photographic,crowson2022vqgan, gafni2022make, rombach2021highresolution,jain2021dreamfields}. 
The use of diffusion models has, in particular, enabled an unprecedented capability in generating high-quality diverse images from natural language input with models such as DALL$\cdot$E2~\cite{ramesh2022hierarchical}, Imagen~\cite{saharia2022photorealistic},  Parti~\cite{yu2022scaling}, Make-A-Scene~\cite{gafni2022make}, Stable Diffusion (SD)~\cite{Rombach_2022_CVPR}, and CogView2~\cite{ding2022cogview2}.

A recent line of work extends models of this kind by tuning the input embeddings to personalize image generation. In particular, some contributions generate images based on a small group of images:
Textual inversion (TI)~\cite{gal2022textual} optimizes the embedding of a new textual token that represents a concept found in a handful of images. DreamBooth~\cite{ruiz2022dreambooth} proposes fine-tuning the \emph{full} image generation model where a unique identifier represents the concept. Both works require 3-5 training images to learn the identity of the target concept. A related line of work enables editing of a given image based on input text or another image~\cite{patashnik2021styleclip, gal2021stylegan, chefer2021targetclip, avrahami2022blended,avrahami2022blendedlatent, bar2022text2live}.

More recently, some have suggested methods to leverage large text-based image generators for image editing. Prompt-to-prompt~\cite{hertz2022prompt} edits the input prompt directly via manipulation of cross-attention maps, and Imagic~\cite{kawar2022imagic} optimizes the corresponding textual prompt and fine-tunes the model such that the image is accurately reconstructed. When only a few images are used for training, though, there is always the inherent risk that a concept can become too similar to the original images.

In contrast, we aim to steer existing diffusion models toward a more general understanding of class via the informative characteristics that a classifier maintains to discriminate between them, while still taking full advantage of the diversity of the underlying generative model. Furthermore, our method is significantly faster than methods such as TI and can effectively utilize an off-the-shelf classifier to refine an image within minutes. %

Manipulating an image using classifier conditioning can also provide a counterfactual explanation for classifiers~\cite{boreiko2022sparse, gat2022latent, augustin2022diffusion}. In that sense, our method might also be used to reveal hidden factors of the classifier used. Since semantic differences are relatively small during each iteration of the image generation, they can easily be detected during the process. %

\section{Method}
\begin{figure}[t]
    \centering
    \includegraphics[width=1\linewidth]{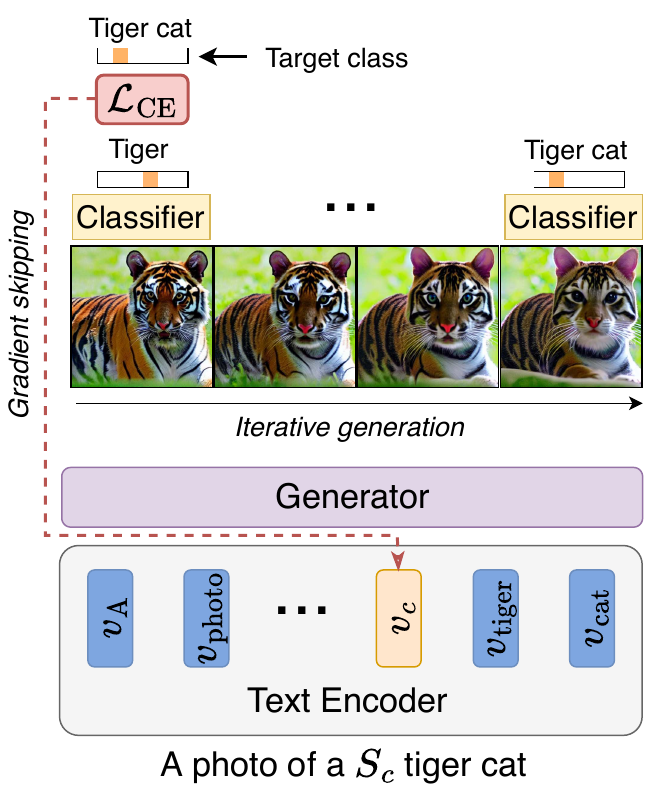}
    \caption{An overview of our method for optimizing a new discriminative token representation ($v_c$) using a pre-trained classifier. For the prompt `A photo of a $S_c$ tiger cat,' we expect the output generated with the class $c$ to be `tiger cat'. The classifier, however, indicates that the class of the generated image is `tiger'. We generate images iteratively and optimize the token representation using cross-entropy. 
    Once $v_c$ has been trained, more images of the target class can be generated by including it in the context of the input text.
    }\label{fig:model}
\end{figure}

We now describe how discriminative token embeddings are learned. We first introduce conditional diffusion models in general, including the more traditional \emph{classifier guidance} (not to be confused with our method), and then describe our conditioning approach and the gradient skipping. An overview of our method is provided in \cref{fig:model}. 

\paragraph{Conditional diffusion models}

Diffusion models~\cite{ho2020denoising, dhariwal2021diffusion} estimate a process that generates data $x\sim p(x)$ from noise. %
During training, an iterative denoising process predicts step-wise added noise %
$x_T \sim \mathcal{N}(0,\text{I})$. More specifically, given an input image (or a latent encoding) $x_0\sim p(x)$, one first produces samples  $x_t = \sqrt{\alpha_t}x_0 + \sqrt{1 - \alpha_t} \cdot \epsilon_t$, with $0 < \alpha_T <  \alpha_{T-1} < \dots < \alpha_0=1$ being hyperparameters and $\epsilon_t \sim \mathcal{N}(0,\text{I})$. One then trains a neural network to predict the added noise $\epsilon_t$ via
an objective function of the form:
\begin{equation}
    \mathbb{E}_{x,\epsilon_t,t}[||\hat \epsilon_\theta(x_t, t) - \epsilon_t||^2_2],
      \label{eq:loss_uncond}
\end{equation}
A conditional denoising process, where each denoising step depends on a conditioning input (e.g., a class identifier or a text prompt $y$), can be defined similarly: 
\begin{equation}
    \mathbb{E}_{x,\epsilon_t,t}[||\hat \epsilon_\theta(x_t, t, y) - \epsilon_t||^2_2],
      \label{eq:loss_cond}
\end{equation}
To condition the diffusion process on a class, gradients calculated with trained classifiers can be used in the denoising process~\cite{dhariwal2021diffusion}. In particular, gradients of text-image matching models, like CLIP~\cite{radford2021learning}, allow text conditioning. Utilizing classifier guidance improves the sample quality and enables a trade-off between sample quality and diversity.

There are two main drawbacks to using classifier guidance within the diffusion process: (i) the classifier must be retrained to deal with noised images as every noisy sample generated along the iterative denoising process must be passed through the classifier, and (ii) the classifier needs to be present throughout the generative process. To deal with this, a \emph{classifier-free} approach has been proposed~\cite{dhariwal2021diffusion}. Instead of relying on gradients from an image classifier, this approach approximates the gradient of an implicit classifier by modeling the difference between conditional, $p_\theta(x|y)$, and unconditional, $p_\theta(x)$, denoising modules. The conditional and unconditional modules are parameterized using the same $\epsilon_\theta(x_t, y)$, and the conditional network becomes unconditional by using an empty sentence, i.e., $\epsilon_\theta(x_t) = \epsilon_\theta(x_t. "")$. The final denoising network is formally expressed as follows:
\begin{equation}
    \bar \epsilon_\theta(x, x_t, y) = (1+w)\epsilon_\theta(y_t, y) - w(\epsilon_\theta(x_t)),
    \label{eq:denoising}
\end{equation}
where $w$ is a hyperparameter determining the strength of the conditioning guidance.

Our method is complementary to both the \emph{classifier-based} and \emph{classifier-free guidance}, and can be used in conjunction with both. 
While our method can be deployed using any diffusion model, we consider Stable Diffusion (SD)~\cite{rombach2021highresolution} where the denoising process is applied not directly to the pixel values of the images, but in the lower dimensional hidden dimensions of a neural network. The latent representation is gotten by using a Variational Autoencoder (VAE) model that maps an image into a latent $z = V(x)$, and decodes from the latent representation back to an image $x\approx D(z) $.

\paragraph{Discriminative Token Embeddings}

Classifiers capture discriminative signals that are useful for discerning between classes. %
In that sense, pretrained classifiers can be seen as experts in different domains. %
For example, a bird classifier can provide a compact source of discriminative details that separate one species from another.%

To avoid relying on a classifier at inference time, and reduce the need to fine-tune the classifier on noised images (as in earlier work), our method fine-tunes a token added to the text encoder's vocabulary that the SD model relies on. Our technique iteratively generates images and refines this added synthetic token (as opposed to a word or subword found in the input language) to associate the generated images with a target class of the pre-trained classifier. The only weights being updated are those of the new class token.

The process starts with a discriminative class token $S_c$ and a generic prompt $p=\text{``A photo of a $S_c$ \textit{class\_name}''}$, where \textit{class\_name} is the name of of the class. We include the class name as part of the prompt to take advantage of existing knowledge in the pre-trained diffusion model (but only update the embedding for $S_c$ during training). By looking at the images, e.g., in \cref{fig:teas} %
it is evident that SD has gained knowledge across various domains, including expertise in generating specific bird species with some limitations. Our approach aims to enhance the precision of the generated image by introducing minor semantic modifications that leverage the model's existing knowledge.

We now describe how we associate the target class's characteristics with the class token's representation. We denote the embedding of the class token $S_c$ as $v_c$, and learn it by utilizing an image classifier $C$. To speed up training, $v_c$ is initialized to be the embedding of a related initializer token. For instance, with a bird classifier, we initialize the embedding to that of the `bird' token in the input encoder. For more general classifiers, such as those trained on ImageNet, we use the indefinite article token `a' as a base for more generalized concepts. %
Training starts by generating an image $x(p)$ conditioned on $p$, including the $v_c$ representation. We feed the resulting image into the classifier and use cross-entropy loss over the classifier labels, i.e.,
\begin{equation}
\min_{v_c}\operatorname{CE}\left( C\left(\psi_C(x(p))\right), \mathds{1}_c \right),
  \label{eq:our_loss}
\end{equation}
where $\psi_C$ transforms the image to align it with the classifier's expected input (e.g., resizing), and $\mathds{1}_c $ is one hot vector of the target class. 
Our method does not rely on a given set of images. Instead, it generates images iteratively, starting with the output from SD, where each optimization step shifts the generated images closer to the target class distribution by updating the class token. A single image can be optimized directly and generally converges in a relatively small number of steps.

\begin{figure}
    \centering
    \includegraphics[width=1\linewidth]{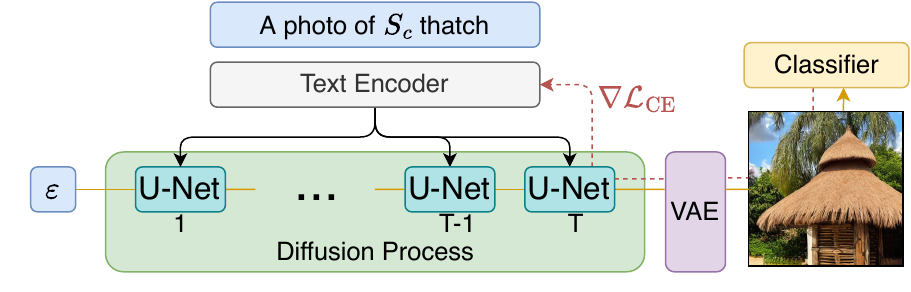}
    \caption{An illustration of the gradient skipping technique (indicated by the red line). During backpropagation, the gradient is propagated only through the final denoising step of the diffusion procedure.
    }\label{fig:backprop}
\end{figure}

\paragraph{Gradient skipping} \cref{fig:backprop} illustrates the generation of a single image using a diffusion process and the flow of the learning signal. Propagating gradients through all diffusion steps requires a significant amount of memory. In our experiments, propagating gradients solely through the final denoising step (i.e., step $T$) produces high-quality images representing the intended class while needing fewer resources. While deeper backpropagation could lead to further enhancements, we do not explore this direction further due to memory constraints. %

\paragraph{Design Choices} 

\begin{figure*}
    \centering
    \includegraphics[width=1\linewidth]{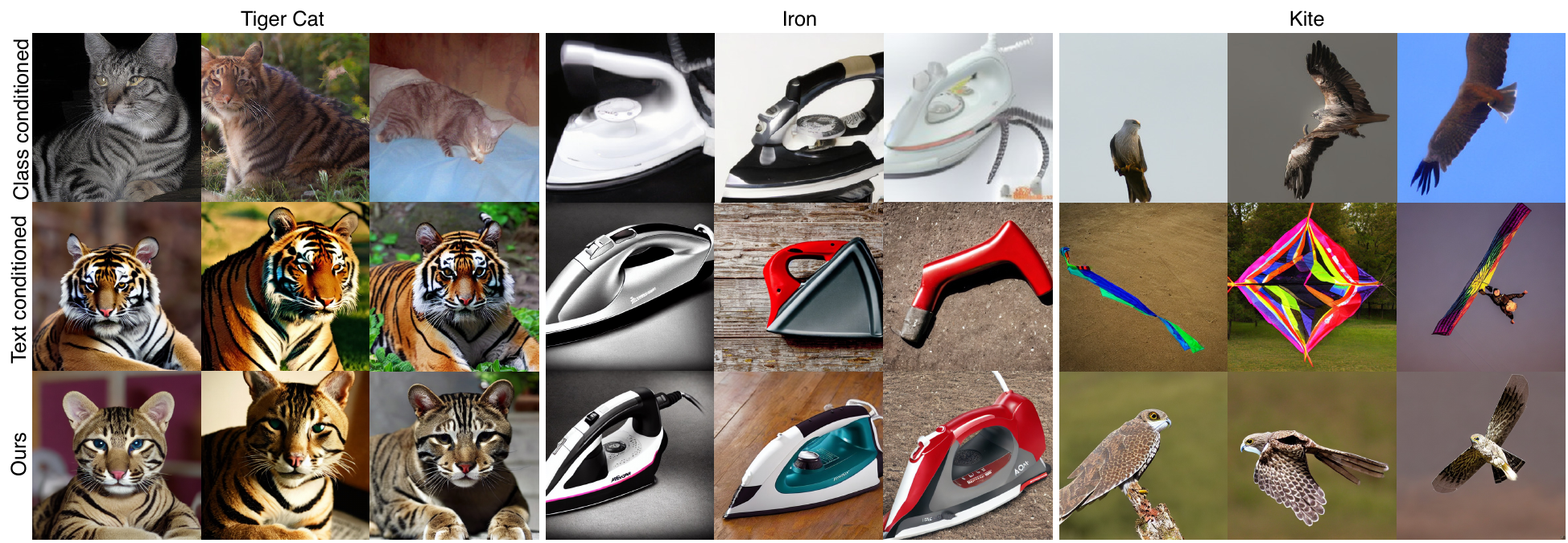}
    \caption{Images using class-conditioned LDM, text-conditioned (SD), and our method with ImageNet classifier guidance.
    }\label{fig:class-condition}

    \centering
    \includegraphics[width=1\linewidth]{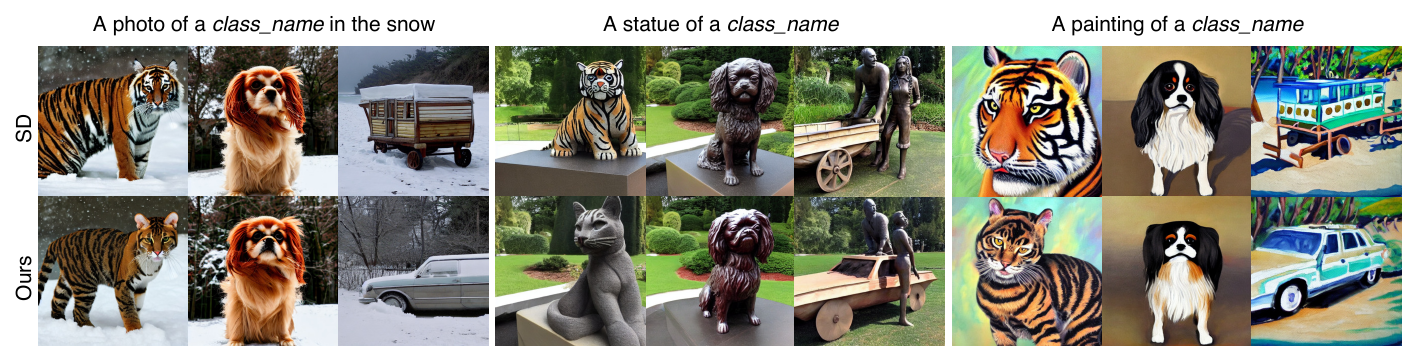}
    \caption{Results with different prompts for three classes: (i) tiger cat, (ii) Japanese spaniel, and (iii) beach wagon.
    }\label{fig:edit_prompts}
\end{figure*}

Our approach involves several design choices. (i) \textit{Batch size}: Our goal is not to refine a single image but to find a broad token representation that can generate new images without incurring extra costs. By generating images with different seeds, we get diverse images. A larger batch size picks up more generic discriminative features, but training takes slightly longer to converge. We set the batch size to 5 after experimenting with values of 1-6. More details and examples of generations are shown in the supplementary.
(ii) \textit{Number of prompts}: Increasing the number of prompts can introduce additional variability. However, we find too much variability during training harmful to convergence. Thus, we limited the number of prompts used in the training phase to two: $p_1 = $``A high-resolution realistic image of a $S_c$ \textit{label}'', and $p_2 = $``A photo of $S_c$ \textit{label}''. 
It is worth noting 
that one can still utilize the discriminative token across various prompts, as shown in \cref{fig:edit_prompts}.
(iii) \textit{Updated tokens}:
Our experiments focus on optimizing only the embedding of $S_c$. While it is possible to update other pre-existing tokens, doing so would modify the model and prevent it e.g. from being used in case of lexical overlap between classes, such as in the case of iron. %
(iv) \textit{Early stopping strategy}:
Please refer to the supplementary for additional details.

\begin{table}
\centering
\begin{tabular}{lllcc}
\toprule
 Dataset & Classifier & Guidance & Top-1 & Top-5 \\
  \midrule 
 ImageNet & ImageNet & - & 70.5 & 89.7\\
ImageNet & ImageNet & ImageNet & \textbf{74.5} & \textbf{92.6}\\
   \midrule 
CUB & CUB & - & 39.7 & 73.3\\
CUB & CUB & CUB & \textbf{57.9} & \textbf{88.6}\\
  \midrule 
  iNat179 & iNat & - & 28.5 & 56.5\\
 iNat179 & iNat & CUB & \textbf{32.8} & \textbf{63.5}\\
 \midrule 
 iNat50 & iNat & - & 14.1 & 28.7\\
iNat50 & iNat & iNat & \textbf{25.8} & \textbf{50.3}\\

\bottomrule
\end{tabular}
\caption{Classification results on four datasets: ImageNet, CUB, and two subsets from iNaturalist21: iNat50, and iNat179. For each dataset, we calculate the accuracy using a classifier (Classifier) on images generated with or without the guidance of another classifier (Guidance). 
For all datasets, accuracy is higher when guidance is used, including in the case (iNat179) where the classifier used (iNat) differs from that used for guidance (CUB). }
\label{tab:results}
\vspace{-0.1cm}
\end{table}

\begin{table}
\small
\begin{center}
\begin{tabularx}{\columnwidth}{r|r|rr|cc|cc}
\toprule
 &  \# Real img.& \multicolumn{2}{c|}{Baseline} &\multicolumn{2}{c|}{SD} & \multicolumn{2}{c}{Ours (CUB)}\\
 & per class  &  T-1 & T-5 & T-1 & T-5 & T-1 & T-5 \\
\midrule
\multirow{4}{*}{\rotatebox[origin=c]{90}{CUB }} &  0  & 0.0 & 0.0 & 37.1 & 67.8 & \textbf{48.6} & \textbf{78.5} \\ %
 & 3  & 1.8  & 5.1 & 52.6 & 84.9 & \textbf{62.5} & \textbf{87.9 }\\ %
 &  9  & 35.1  & 72.3 & 68.3 & 92.7 & \textbf{76.3} & \textbf{95.1} \\ %
 & 15  & 71.5  & 94.1  & 77.3 & 95.8 & \textbf{81.4} & \textbf{96.4} \\ %
\midrule
\multirow{4}{*}{\rotatebox[origin=c]{90}{iNat179}}  & 0 & 0.0 & 0.0 & 22.1 & 47.7 & \textbf{28.1} & \textbf{55.2} \\ %
 & 3 & 1.3 & 5.7 & 31.6 & 61.1 & \textbf{37.1} & \textbf{66.9}\\ %
 &  9 & 8.0 & 27.3 & 43.0 & 74.3 & \textbf{49.5} & \textbf{78.3} \\ %
 & 15 & 28.7 & 62.9 & 49.8 & 81.4 & \textbf{58.3} & \textbf{84.1} \\ %

\bottomrule
\end{tabularx}
\vspace{-0.1cm}
\end{center}
\caption{Results for classifiers trained with 100 generated and 0-15 real images (\# Real img.) from each CUB class, and in the overlap of 179 species in iNat. For the baseline column, we use only real images. Results with our method use \textbf{CUB guidance} and improve performance even for the different iNat179 dataset. }

\label{tab:genresults}
\end{table}

\begin{table}
\small
\centering
\begin{tabular}{lc}
\toprule
 Method & FID $\downarrow$  \\
\midrule
Text-conditioned (SD) & 23.0 \\
\midrule
\multicolumn{2}{c}{\textit{Class Conditioned Methods}}\\
\midrule
Class-conditioned~\cite{rombach2022high} & 47.6 \\
Ours & \textbf{22.4} \\
\bottomrule
\end{tabular}
\caption{FID scores for generated ImageNet classes.}
\label{tab:classfid}
\vspace{-0.2cm}
\end{table}

\begin{figure*}[t]
        \includegraphics[width=1\linewidth]{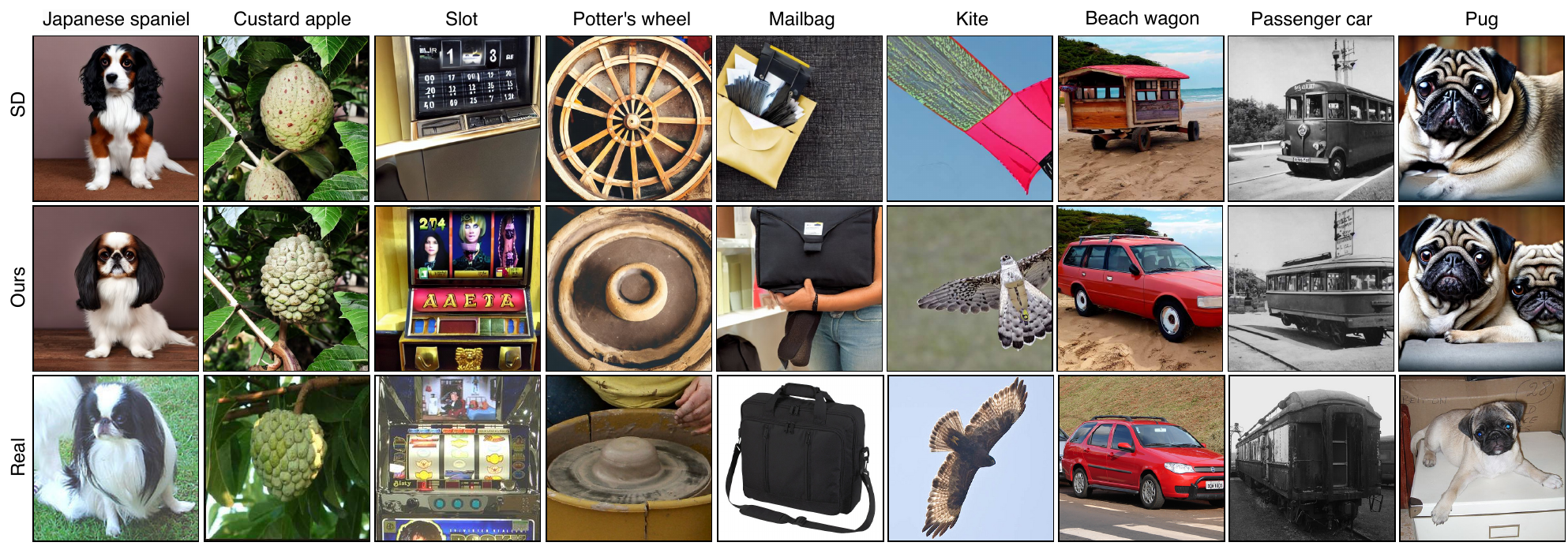}\vspace{-5pt}
    \caption{Images generated based on ImageNet classes, using SD or our method. Real images are shown for comparison. 
    }\label{fig:inet-qual}\vspace{-15pt}
\end{figure*}

\begin{figure}[t]
    \centering
    ~~~~
    \begin{subfigure}[b]{0.11\textwidth}
    ~~~~~~ (a)
    \end{subfigure}
    \begin{subfigure}[b]{0.11\textwidth}
   ~~~~~~ (b)
    \end{subfigure}
    \begin{subfigure}[b]{0.11\textwidth}
~~~~~~(c)
    \end{subfigure}
    \begin{subfigure}[b]{0.11\textwidth}
~~~~~~(d)
    \end{subfigure}

    \begin{subfigure}[b]{0.015\textwidth}
    \adjustbox{varwidth=1cm,raise=1.05cm}{{\rotatebox[origin=c]{90} {SD}  }}
     \end{subfigure}
    \begin{subfigure}[b]{0.11\textwidth}
         \centering
         
         \includegraphics[width=\linewidth]{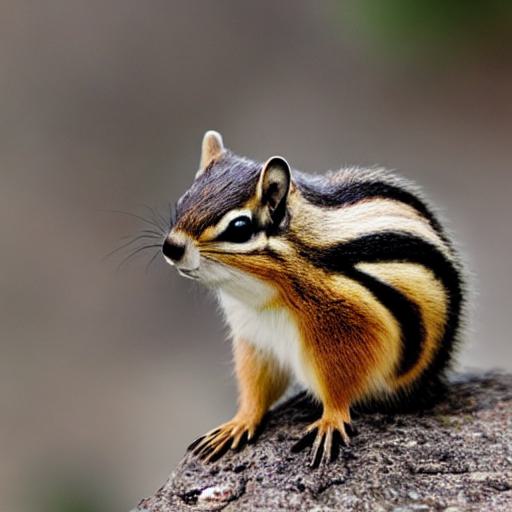}
     \end{subfigure}
     \begin{subfigure}[b]{0.11\textwidth}
         \centering
         \includegraphics[width=\linewidth]{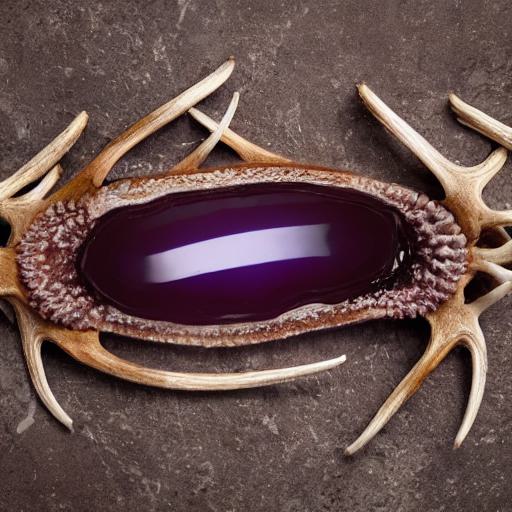}
     \end{subfigure}
     \begin{subfigure}[b]{0.11\textwidth}
         \centering
         \includegraphics[width=\linewidth]{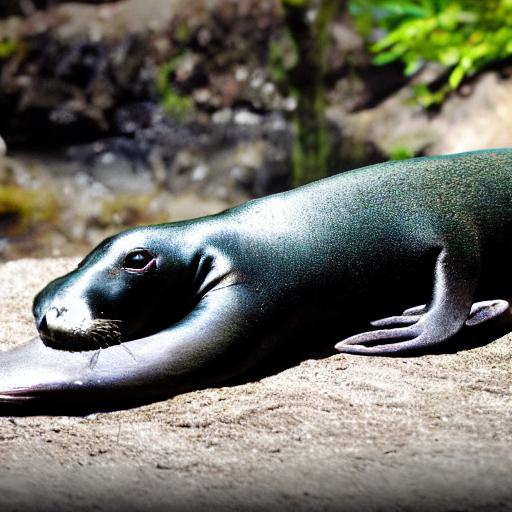}
     \end{subfigure} 
     \begin{subfigure}[b]{0.11\textwidth}
         \centering
         \includegraphics[width=\linewidth]{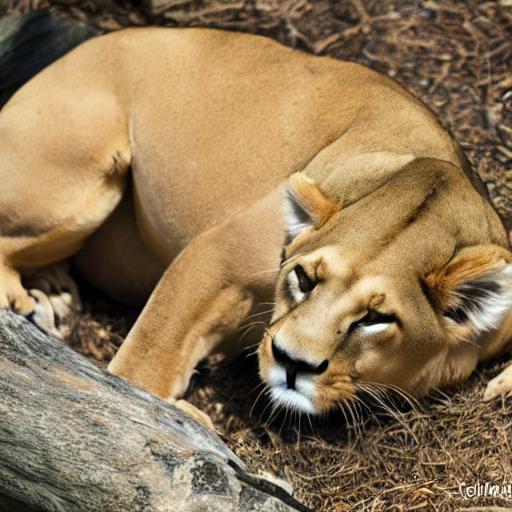} \\
     \end{subfigure} 
     
         \begin{subfigure}[b]{0.015\textwidth}
    \adjustbox{varwidth=1cm,raise=1.05cm}{{\rotatebox[origin=c]{90} {Ours}}}
     \end{subfigure}
    \begin{subfigure}[b]{0.11\textwidth}
         \centering
         
         \includegraphics[width=\linewidth]{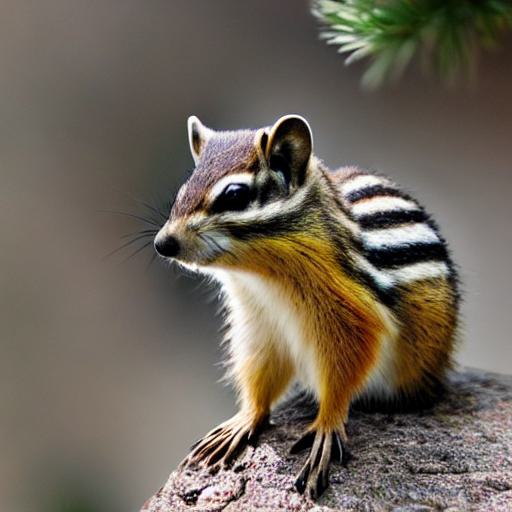}
     \end{subfigure}
     \begin{subfigure}[b]{0.11\textwidth}
         \centering
         \includegraphics[width=\linewidth]{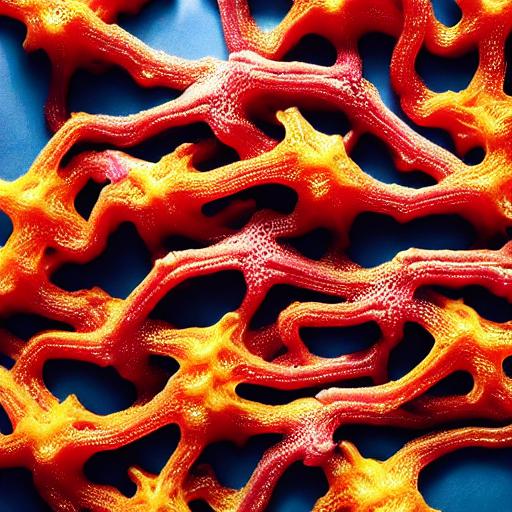}
     \end{subfigure}
     \begin{subfigure}[b]{0.11\textwidth}
         \centering
         \includegraphics[width=\linewidth]{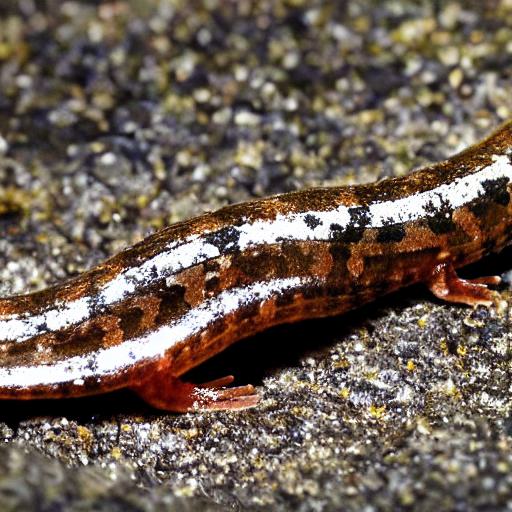}
     \end{subfigure} 
     \begin{subfigure}[b]{0.11\textwidth}
         \centering
         \includegraphics[width=\linewidth]{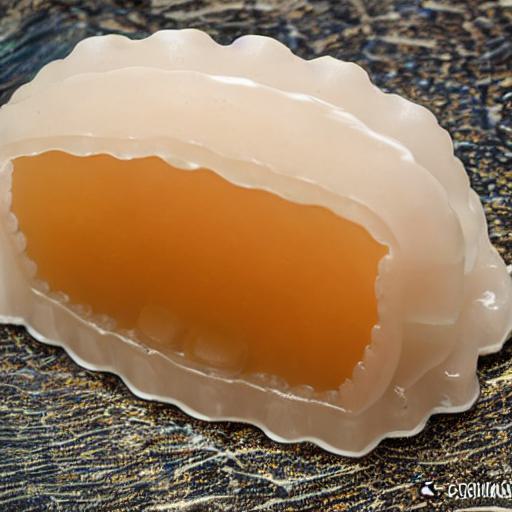} \\
     \end{subfigure} 
     
              \begin{subfigure}[b]{0.015\textwidth}
    \adjustbox{varwidth=1cm,raise=1.05cm}{{\rotatebox[origin=c]{90} {Real}}}
     \end{subfigure}
    \begin{subfigure}[b]{0.11\textwidth}
         \centering
         
         \includegraphics[width=\linewidth, height=\linewidth]{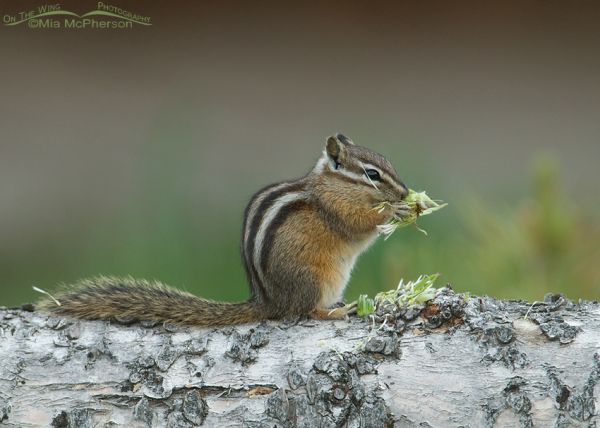}
     \end{subfigure}
     \begin{subfigure}[b]{0.11\textwidth}
         \centering
         \includegraphics[width=\linewidth, height=\linewidth]{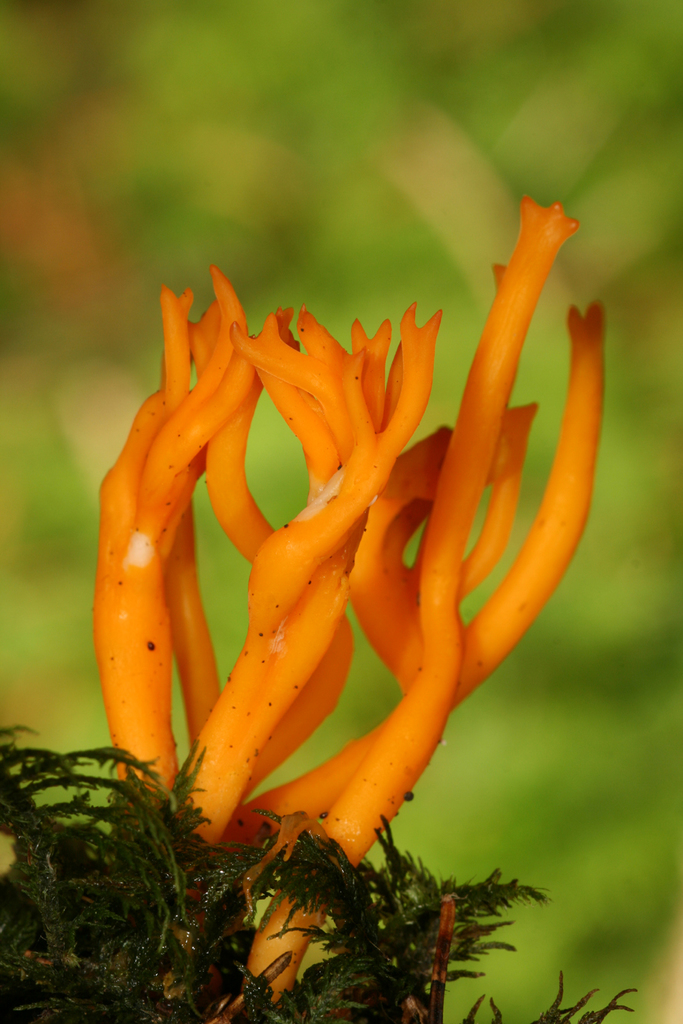}
     \end{subfigure}
     \begin{subfigure}[b]{0.11\textwidth}
         \centering
         \includegraphics[width=\linewidth, height=\linewidth]{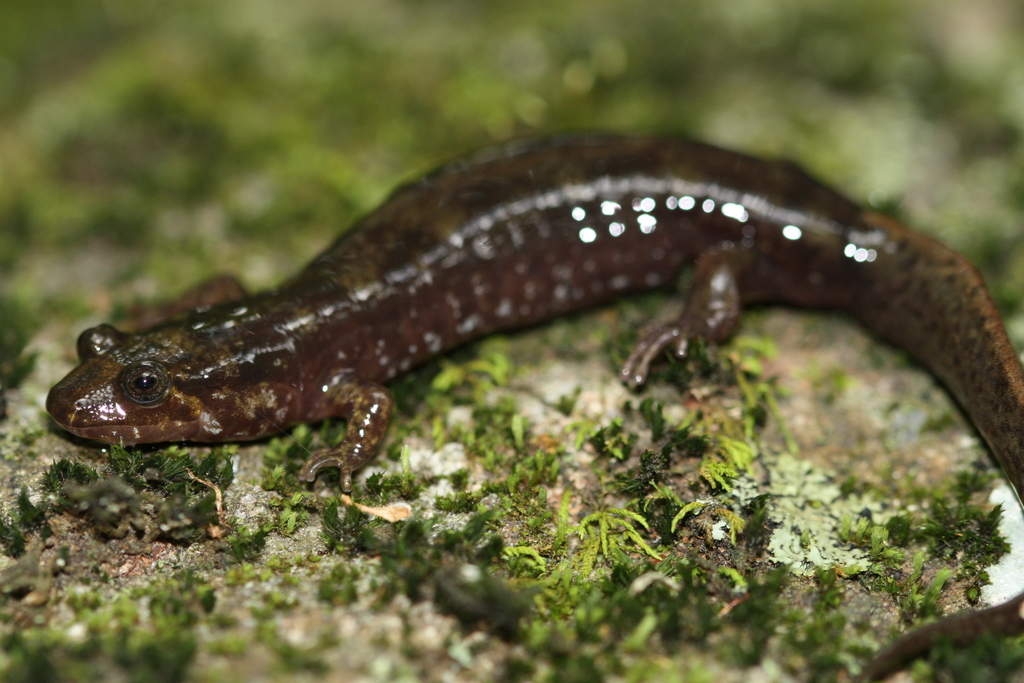}
     \end{subfigure} 
     \begin{subfigure}[b]{0.11\textwidth}
         \centering
         \includegraphics[width=\linewidth, height=\linewidth]{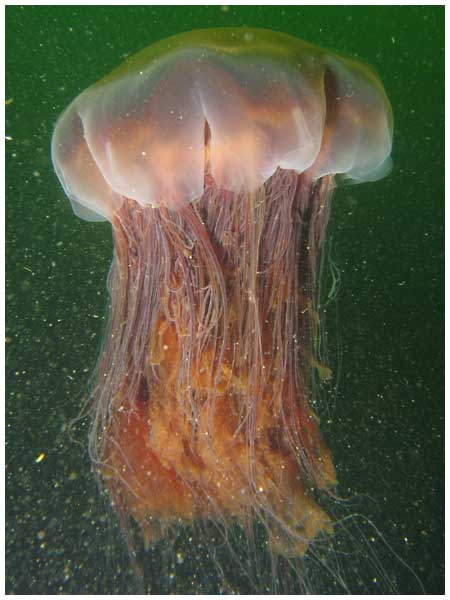} \\
     \end{subfigure}

    \begin{subfigure}[b]{0.11\textwidth}
~~~~~~~~~~(e)
    \end{subfigure}
    \begin{subfigure}[b]{0.11\textwidth}
~~~~~~~~~~(f)
    \end{subfigure}
     \begin{subfigure}[b]{0.11\textwidth}
    ~~~~~~~~~~ (g)
    \end{subfigure}
    \begin{subfigure}[b]{0.11\textwidth}
    ~~~~~~~~~~(h)
    \end{subfigure}

    \begin{subfigure}[b]{0.015\textwidth}
    \adjustbox{varwidth=1cm,raise=1.05cm}{{\rotatebox[origin=c]{90} {SD}}}
     \end{subfigure}
    \begin{subfigure}[b]{0.11\textwidth}
         \centering
         
         \includegraphics[width=\linewidth]{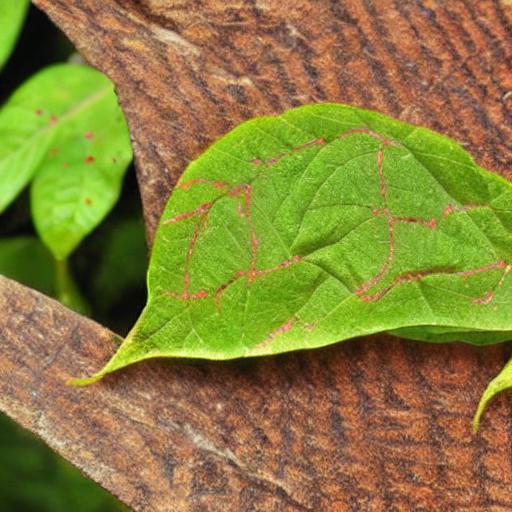}
     \end{subfigure}
     \begin{subfigure}[b]{0.11\textwidth}
         \centering
         \includegraphics[width=\linewidth]{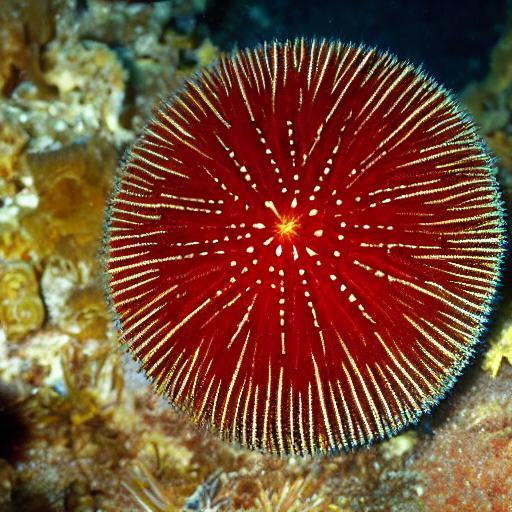}
     \end{subfigure}
     \begin{subfigure}[b]{0.11\textwidth}
         \centering
         \includegraphics[width=\linewidth]{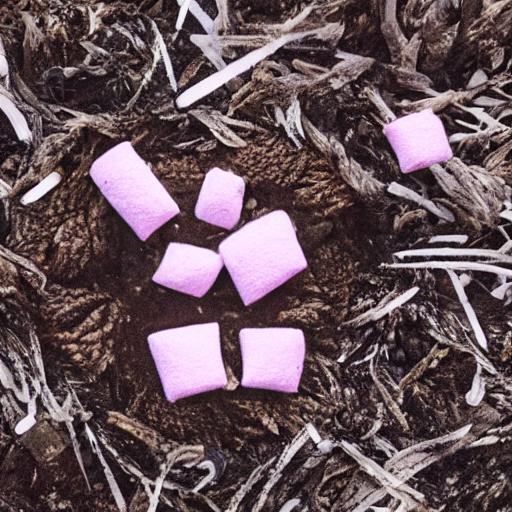}
     \end{subfigure} 
     \begin{subfigure}[b]{0.11\textwidth}
         \centering
         \includegraphics[width=\linewidth]{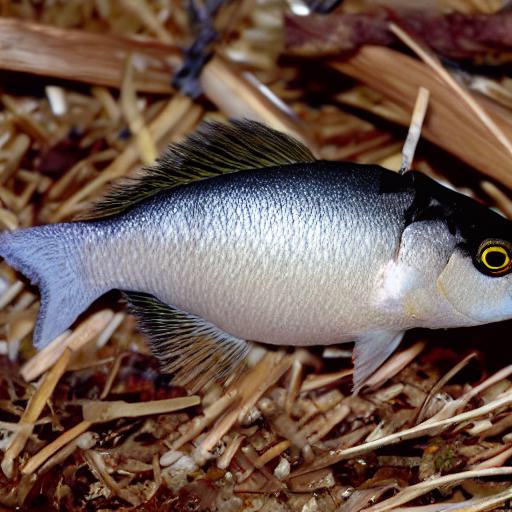} \\
     \end{subfigure} 
     
         \begin{subfigure}[b]{0.015\textwidth}
    \adjustbox{varwidth=1cm,raise=1.05cm}{{\rotatebox[origin=c]{90} {Ours}}}
     \end{subfigure}
    \begin{subfigure}[b]{0.11\textwidth}
         \centering
         
         \includegraphics[width=\linewidth]{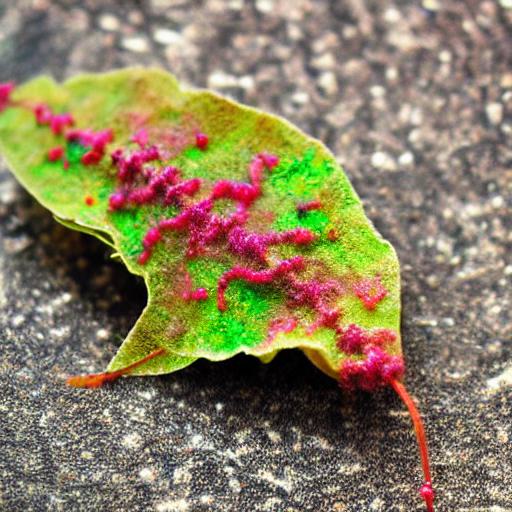}
     \end{subfigure}
     \begin{subfigure}[b]{0.11\textwidth}
         \centering
         \includegraphics[width=\linewidth]{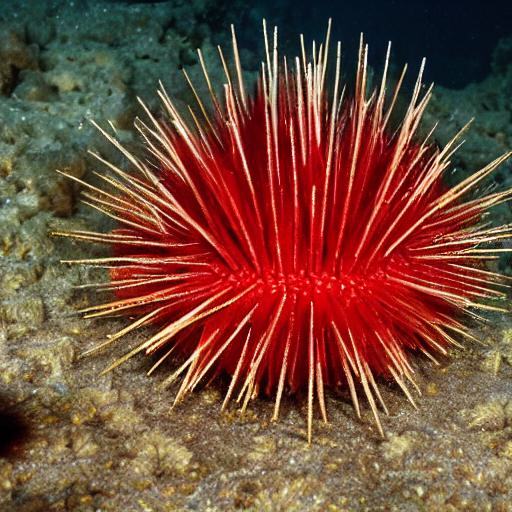}
     \end{subfigure}
     \begin{subfigure}[b]{0.11\textwidth}
         \centering
         \includegraphics[width=\linewidth]{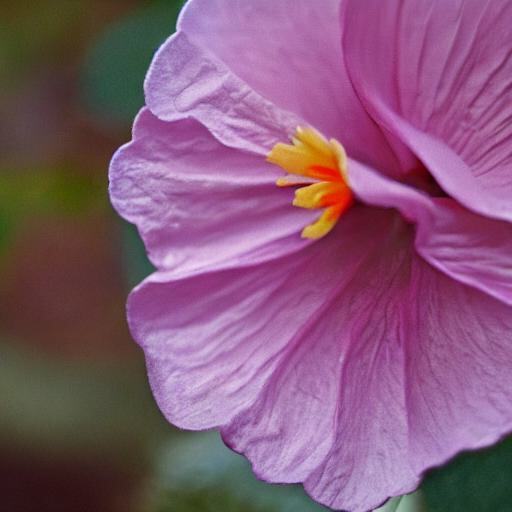}
     \end{subfigure} 
     \begin{subfigure}[b]{0.11\textwidth}
         \centering
         \includegraphics[width=\linewidth]{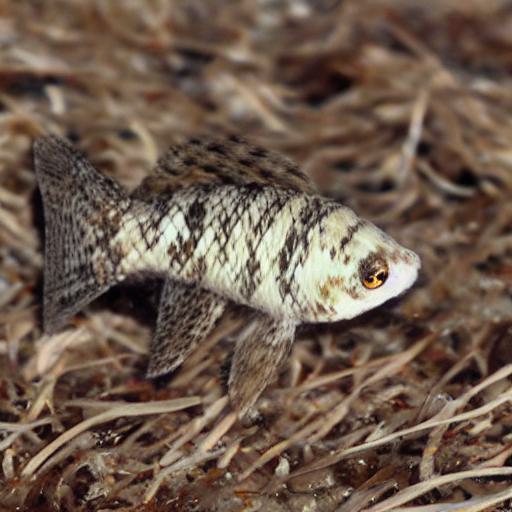} \\
     \end{subfigure} 
     
              \begin{subfigure}[b]{0.015\textwidth}
    \adjustbox{varwidth=1cm,raise=1.05cm}{{\rotatebox[origin=c]{90} {Real}}}
     \end{subfigure}
    \begin{subfigure}[b]{0.11\textwidth}
         \centering
         
         \includegraphics[width=\linewidth, height=\linewidth]{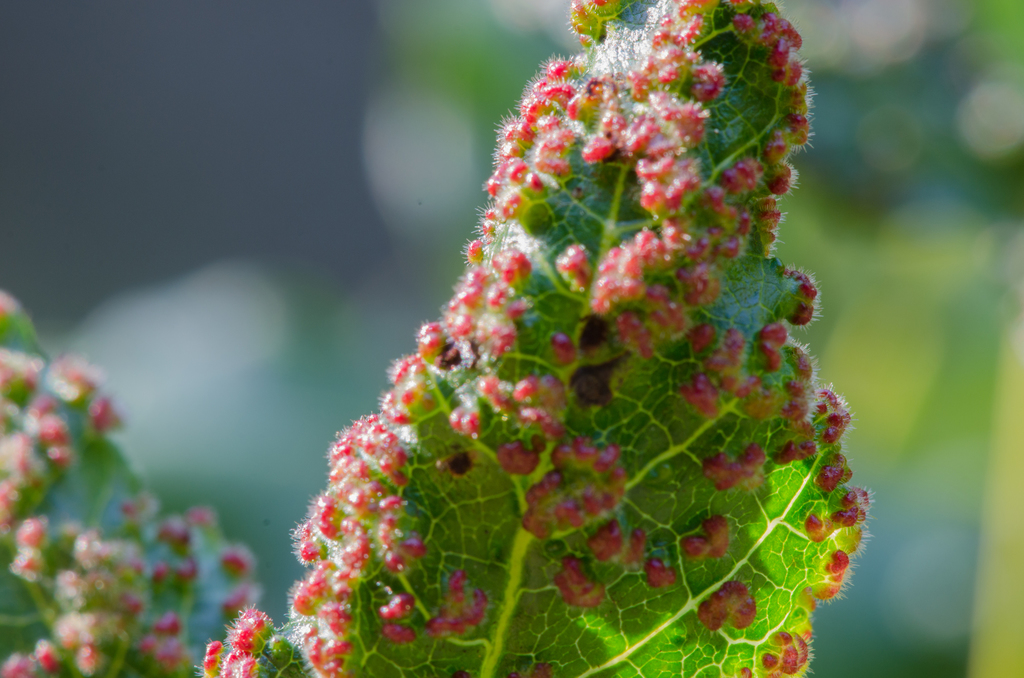}
     \end{subfigure}
     \begin{subfigure}[b]{0.11\textwidth}
         \centering
         \includegraphics[width=\linewidth, height=\linewidth]{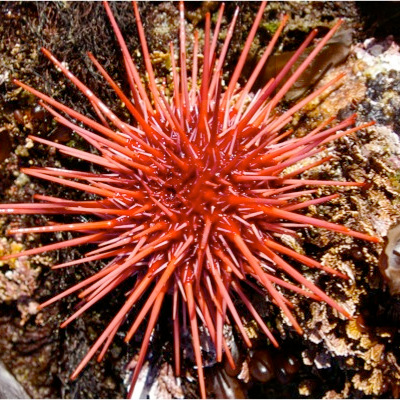}
     \end{subfigure}
     \begin{subfigure}[b]{0.11\textwidth}
         \centering
         \includegraphics[width=\linewidth, height=\linewidth]{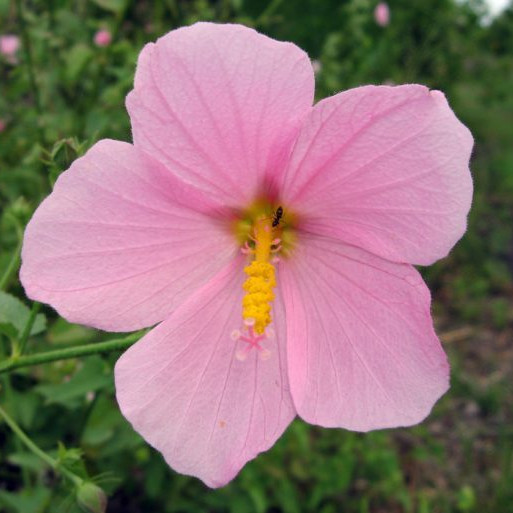}
     \end{subfigure} 
     \begin{subfigure}[b]{0.11\textwidth}
         \centering
         \includegraphics[width=\linewidth, height=\linewidth]{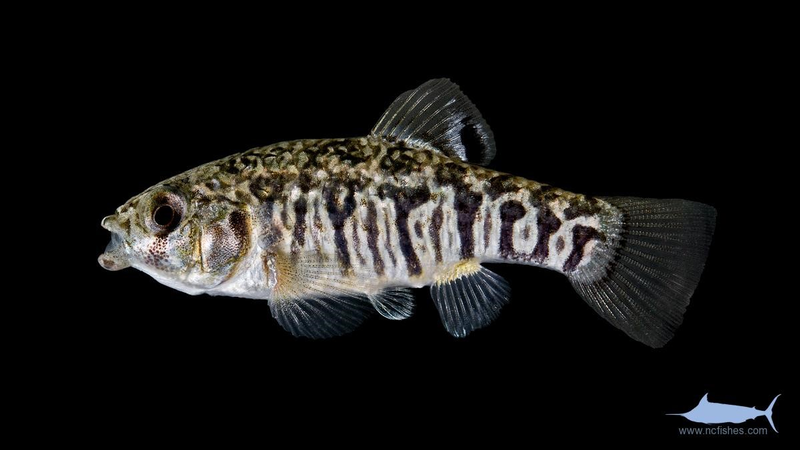} \\
     \end{subfigure} \vspace{-20pt}
    
        \caption{A selection of images based on iNat classes generated with Stable Diffusion (SD) and our method. A real image is shown for comparison. (a). Yellow pine 
    chipmunk, (b). Jelly antler, (c). Salamander, (d). Pacific lions mane jelly, (e). Leaf mite, (f). Red sea urchin, (g). Seashore mallow, (h). Sheepshead minnow.  \label{fig:inat}}\vspace{-20pt}

\end{figure}

\section{Results}

Quantitatively, we evaluate the ability of our method to conform to the input class and to generate high-fidelity images. For the former, we consider the following:
(i) taking pre-trained classifiers and assessing the classification accuracy of generated images, and (ii) evaluating the accuracy of classifiers trained using generated and real images. This assesses the generated images in two complementary ways. If the pre-trained classifier correctly classifies generated images, then they capture features from the correct class. If the generated images can be used to improve a classifier's performance, then they capture additional discriminative features which can improve classification accuracy. To evaluate the quality of generated images, we consider the commonly used FID score~\cite{NIPS2017_8a1d6947}. 
Qualitatively, we demonstrate the effectiveness of our approach in adding fine-grained details and resolving problematic cases of ambiguity. Please refer to the supplementary for a full description of the experimental setup including implementation and training details.

\paragraph{Datasets} 
For \textit{fine-grained} categories, such as those of birds and animal species, we use the CUB dataset~\cite{WahCUB_200_2011} and the iNaturalist 2021-mini (iNat from here on) dataset~\cite{su2021semi_iNat}. We also consider two subsets of iNat: (a) iNat179, a subset that overlaps with CUB in 179 out of the 200 labels and allows us to classify images generated when utilizing guidance from a different classifier (in particular, one trained on a different dataset), (b) iNat50, a sample of 50 species randomly selected from each of the 11 supercategories. In the \textit{course-grained} setting, we consider the ImageNet dataset~\cite{deng2009imagenet}.

\subsection{Quantitative evaluation}

\paragraph{Evaluation using pre-trained classifiers}
\label{sec:result_classifier}

We first evaluate the generated images with classifiers trained on real data. We generate 100 images for each class and calculate the accuracy of each method: one employing our class token guidance and another with only SD. In \cref{tab:results}, we show that for ImageNet, which mainly consists of classes at a coarse level of granularity, the vanilla SD can generate most classes accurately (70.5\%). By utilizing class token guidance, we get better results in complex cases, such as those with ambiguity, resulting in an improved accuracy of 74.5\%. For our method and SD, the seed and textual context were held constant, so the images correspond to each other, with the differences being due to the use of the token.

We next assess fine-grained classes. Testing the model on the images generated with labels found in the CUB dataset, the accuracy of SD-generated images is only 39.7\%. This emphasizes the inherent limitations of the SD model in generating highly detailed and specific categories, such as bird species. Our approach adds the fine details necessary to improve accuracy to 57.9\%.
We also assess the accuracy of the generated images using the iNat classifier (trained on 10k species) on the same images %
generated with the guidance of the CUB classifier. %
Our approach yields a noteworthy improvement in performance from 28.5\% to 32.8\%, indicating its potential to enhance fine-grained classes beyond the specific classifier-selected characteristics. Finally, we look at a diverse set of 50 classes from the iNat dataset. SD only generates images that are classified accurately 14.1\% of the time.
Our method significantly improves accuracy to 25.8\%, although there is still room for further improvement.%

\paragraph{Evaluation by training classifiers}
In \cref{tab:genresults}, we train classifiers on 100 generated images and 0-15 real images. When we incorporate the classifier-guided generated images, the real evaluation accuracy is better than when augmenting only with SD-generated images. With only 9 real images per class, we already reach 76.3\% accuracy compared to 35.1\% with no generated images and 68.3\% with SD-only augmentations. A high accuracy indicates that generated images capture a large part of the image distribution necessary to classify images correctly. Our findings also show that utilizing the CUB classifier for generating images can enhance performance when evaluated in the iNat179 setting. Specifically, incorporating our proposed image augmentation method improves accuracy, reaching 58.3\% with only 15 real images, compared to the significantly lower accuracy of 49.8\% and 28\% with SD or when no augmentation is applied. These results indicate that our approach shows potential for augmenting data in low-resource settings by transferring knowledge from diverse classifiers. %

\paragraph{FID evaluation}
Class-conditioned image generation models may enjoy the benefit of eliminating ambiguity in generated images. However, datasets used to train these models are limited only to specific classes, and so do not capture the wide variety of images depicting free-form text. 
In \cref{tab:classfid}, we utilize the FID~\cite{NIPS2017_8a1d6947} to assess the quality of generated images with respect to the real datasets. Our evaluation shows that the text-conditioned method generates higher quality images compared to a prior class-conditioned method (23.0 vs. 47.6). The text-conditioned method of SD, on the other hand, is limited by ambiguity issues and has difficulty in depicting fine details. Our proposed method provides a balance between generating high-quality images accurately and avoiding ambiguity issues.

\subsection{Qualitative assessment}

\begin{figure*}
    \centering
    \includegraphics[width=1\linewidth]{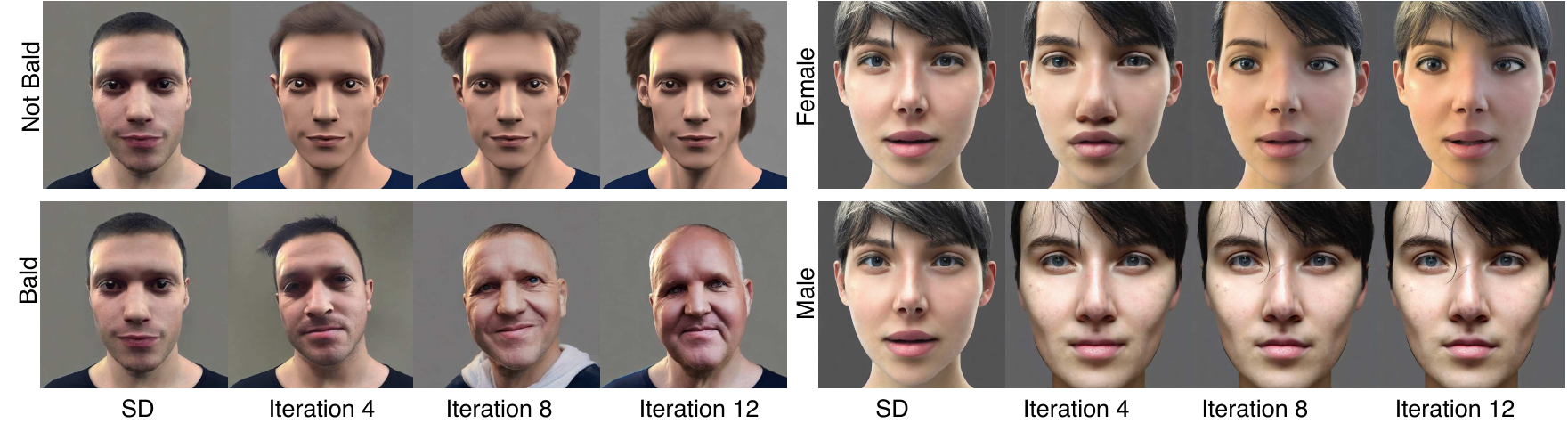}\vspace{-10pt}
    \caption{Demonstration of using discriminative tokens for gender and bald attributes, `SD' shows the initial generation.%
    }\label{fig:celeba-qual}\vspace{-15pt}
\end{figure*}

\begin{figure}
    \centering
    \includegraphics[width=1\linewidth]{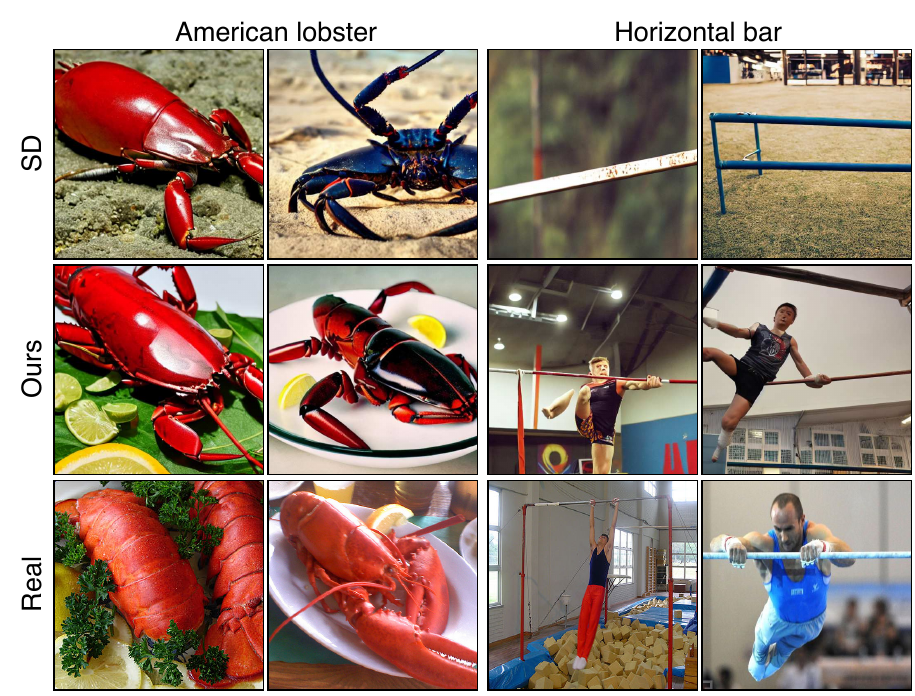}\vspace{-5pt}
    \caption{Examples of revealed features from the training data when using an ImageNet classifier for guidance.
    }\vspace{-15pt}\label{fig:inet-bias}
\end{figure}

In \cref{fig:inet-qual}, we show various generated samples based on ImageNet classes. From left to right, our method reinforces distinctive features of the dog species, in particular the face.%
The distinct characteristics of the custard apple and slot images are highlighted with our method. For the mailbag, the `mail' term appears to confuse SD to generate a mail-related image rather than a mailbag. More ambiguities arise from the term wheel in potter's wheel and the kite class. %
 In some cases, we only partially resolve ambiguity. For instance, beach wagon cars still appear on the beach, and it is still possible to see a kite bird that resembles a kite. Another interesting case is when the method adds another instance of the target class object, as is the case in the pug image. %

In \cref{fig:inat}, we show images generated using labels from the iNat dataset. We sample 50 species from each and compare them using SD and our method. Our method corrects for attributes such as patterns (e.g., (a), and (h)), anatomical issues (e.g., (b), (c), (f)), and resolves lexical ambiguity (e.g., (b), (d), (g)).

In \cref{fig:class-condition}, we show that only class-conditioned images appear less natural despite being highly relevant to the class. Our approach allows us to benefit from the advantages of both worlds, producing high-quality images that are both precise and devoid of ambiguity.

Another advantage of our approach over simple class conditioning~\cite{rombach2022high} is the flexibility to use trained tokens with various prompts. In \cref{fig:edit_prompts}, we demonstrate that our discriminative tokens can be employed in different prompts, resulting in minimal semantic changes that primarily affect the object of interest that is relevant to the class.

\paragraph{Face attributes} Our method is not only capable of enhancing objects and animals. %
For example, we demonstrate that a classifier based on CelebFaces attributes~\cite{liu2015deep} can be utilized to learn a token representing a facial attribute. We generate a facial image using the prompt ``An image of a $S_c$ person's face.''. We optimized $S_c$ using a classifier consisting of six convolutional layers followed by two fully connected layers. In \cref{fig:celeba-qual}, we present our results obtained by training with the guidance of baldness and gender attributes. During training, we observed that the hair feature is more dominant for the `Not bald' class. Interestingly, for the `bald' class, the generated image depicts old men, and their identity is lost. This finding suggests that age may be a hidden factor in the training data. We further explore the idea of revealing concepts in the training data in the next paragraph.

\paragraph{Classifier inversion}
\label{sec:limitations}
Our method can inverse the action of a classifier without access to its trained data. For example, we often observe changes in the background when optimizing for an object's class and as \cref{fig:inet-bias} shows, applying our method with an ImageNet-trained classifier results in an image of a lobster on a plate given the phrase `American lobster.' Another example is the `horizontal bar' class, for which our method predominantly generates images containing athletes and a gym environment. We manually assessed ImageNet's training data by classifying 100 images from the `American lobster' and `horizontal bar' classes and determining whether they exhibit these features in the training data. For the `American lobster' class, 55\% of the images featured a plate, and the lobster in an edible form, and for the `horizontal bar' class, 95\% of the images included an athlete performer. In \cref{fig:inet-bias}, we also present some instances from the training data that illustrate these characteristics. Nevertheless, interpreting the results needs to be done with caution. It is possible that this bias toward a certain type of image reflects one local minimum in our optimization process. %

\section{Conclusion}
This paper introduced a ``plug-and-play'' approach for rapidly fine-tuning text-to-image diffusion models using a discriminative signal. Our approach trains a new token without additional images, enhancing fine-grained details for classifiers pre-trained on datasets such as CUB and iNat and resolving lexical ambiguity. We have also demonstrated how our method can be used to distill generative image models to supplement datasets lacking imagery, edit faces based on attributes classifier, and analyze hidden factors in the training data. Going forward, we aim to extend our approach to other model types beyond classification.

{\small
\bibliographystyle{ieee_fullname}
\bibliography{egbib}
}

\end{document}